\documentclass[letterpaper, 10pt, conference]{ieeeconf}
\IEEEoverridecommandlockouts
   \usepackage[pdftex]{graphicx}
 
  \graphicspath{{../pdf/}{../jpeg/}}
  \DeclareGraphicsExtensions{.pdf,.jpeg,.png}
  
  \usepackage{amssymb} 
  \usepackage{graphicx}
  \usepackage{algorithm}
  \usepackage{algorithmicx}
  \usepackage{algpseudocode}
  \usepackage{dblfloatfix}
  \usepackage{siunitx}
  \DeclareSIUnit\pixel{px}
\usepackage[natural]{xcolor}

\usepackage[cmex10]{amsmath}
\usepackage[tight,footnotesize]{subfigure}
\usepackage{url}

\begin{document}
%%%%%%%%%%%%%%%%%%%%%%title%%%%%%%%%%%%%%%%%%%%%%%%%%%%%%%%%%%%%%%%%

\title{\LARGE \bf Early Start Intention Detection of Cyclists Using Motion History Images and a Deep Residual Network }

\author{Stefan Zernetsch, Viktor Kress, Bernhard Sick and Konrad Doll
	% <-this % stops a space
	%\thanks{This work was not supported by any organization}% <-this % stops a space
	\thanks{S. Zernetsch, V. Kress and K. Doll are with the Faculty of Engineering,
		University of Applied Sciences Aschaffenburg, Aschaffenburg, Germany
		{\tt\footnotesize stefan.zernetsch@h-ab.de, viktor.kress@h-ab.de, konrad.doll@h-ab.de}}
	\thanks{B. Sick is with the Intelligent Embedded Systems Lab, University of Kassel,
		Kassel, Germany
		{\tt\footnotesize bsick@uni-kassel.de}}%
}
%%%%%%%%%%%%%%%%%%%%%%%%%%%%%%%%%%%%%%%%%%%%%%%%%%%%%%%%%%%%%%%%%%%%

% make the title area
\maketitle

%%%%%%%%%%%%%%%%%%%%%%abstract%%%%%%%%%%%%%%%%%%%%%%%%%%%%%%%%%%%%%
%\include{sections/abstract}
\begin{abstract}
	
In this article, we present a novel approach to detect starting motions of cyclists in real world traffic scenarios based on Motion History Images (MHIs). The method uses a deep Convolutional Neural Network (CNN) with a residual network architecture (ResNet), which is commonly used in image classification and detection tasks. By combining MHIs with a ResNet classifier and performing a frame by frame classification of the MHIs, we are able to detect starting motions in image sequences. The detection is performed using a wide angle stereo camera system at an urban intersection. We compare our algorithm to an existing method to detect movement transitions of pedestrians that uses MHIs in combination with a Histograms of Oriented Gradients (HOG) like descriptor and a Support Vector Machine (SVM), which we adapted to cyclists.
To train and evaluate the methods a dataset containing MHIs of 394 cyclist starting motions was created. 
The results show that both methods can be used to detect starting motions of cyclists. Using the SVM approach, we were able to safely detect starting motions \SI{0.506}{\second} on average after the bicycle starts moving with an \textit{F}\textsubscript{1}-score of 97.7\%. The ResNet approach achieved an \textit{F}\textsubscript{1}-score of 100\% at an average detection time of \SI{0.144}{\second}. The ResNet approach outperformed the SVM approach in both robustness against false positive detections and detection time.

\end{abstract}

%%%%%%%%%%%%%%%%%%%%%%%%%%%%%%%%%%%%%%%%%%%%%%%%%%%%%%%%%%%%%%%%%%%%

% no keywords

% For peer review papers, you can put extra information on the cover
% page as needed:
% \ifCLASSOPTIONpeerreview
% \begin{center} \bfseries EDICS Category: 3-BBND \end{center}
% \fi
%
% For peerreview papers, this IEEEtran command inserts a page break and
% creates the second title. It will be ignored for other modes.
\IEEEpeerreviewmaketitle

% no \IEEEPARstart
%This demo file is intended to serve as a ``starter file''
%for IEEE conference papers produced under \LaTeX\ using
%IEEEtran.cls version 1.7 and later.
% You must have at least 2 lines in the paragraph with the drop letter
% (should never be an issue)

% Note that IEEE does not put floats in the very first column - or typically
% anywhere on the first page for that matter. Also, in-text middle ("here")
% positioning is not used. Most IEEE journals/conferences use top floats
% exclusively. Note that, LaTeX2e, unlike IEEE journals/conferences, places
% footnotes above bottom floats. This can be corrected via the \fnbelowfloat
% command of the stfloats package.

\section{\large Introduction}
\label{sec_introduction}

\subsection{Motivation}

Vulnerable road users (VRUs) such as pedestrians and cyclists are an essential part of today's urban traffic. As reported in \cite{WHO.2015}, they are exposed to a considerable danger. 49\% of all persons killed in road accidents worldwide are pedestrians, cyclists, and motorcyclists. Therefore, the protection of VRUs needs to be improved by Advanced Driver Assistance Systems, automated driving functions and infrastructure-based systems. By forecasting the trajectory of VRUs potentially dangerous situations can be detected earlier, e.g., emergency braking can be initiated more rapidly. A mere fraction of a second reduces the risk of serious injuries considerably \cite{Keller.2011}. In addition, trajectory forecasting of VRUs benefits from an early and reliable detection of starting motions, as shown in \cite{Bieshaar.2017}. While pedestrian movement detection has been analyzed before, e.g., in \cite{KoehlerMag} and \cite{Quintero.2017}, cyclists have gained less attention. The proposed system is dedicated to detect cyclist starting motions using infrastructure based sensors which can be part of future intelligent network traffic systems. However, an incorporation into a moving vehicle is possible.

\subsection{Related Work}

Research in the field of intention detection of pedestrians, i.e., detection of basic movements such as \textit{starting} or \textit{turning} and trajectory forecasting, has become more active over the past few years. Keller and Gavrila \cite{Keller.2014} studied the scenario of a stopping or continuing pedestrian at a curbside. They were able to predict a pedestrian's path from a moving vehicle by use of features gained from image-based dense optical flow. In addition, they were able to early detect stopping intentions.

In \cite{KoehlerMag}, K\"ohler et al. detected a pedestrian's intention to enter a traffic lane with help of an SVM. Therefore, a Motion Contour image based Histograms of Oriented Gradients descriptor (MCHOG) was introduced. The motion contours included in MHIs were generated by means of a stationary camera under laboratory conditions and at a real world public intersection. Overall, an accuracy of 99\% for starting detection was reached within the first step of the pedestrian. In \cite{Koehler2015}, this method was transformed for usage in a moving vehicle and extended by stopping and bending in intentions.

Quintero et al. \cite{Quintero.2015} used Balanced Gaussian Process Dynamical Models and a na\"ive-Bayes classifier for intention and pose prediction of pedestrians based on 3D joint positions. This approach was extended by a Hidden Markov Model in \cite{Quintero.2017}. They reached an accuracy of 95.13\% for intention detection and were able to detect starting motions \SI{0.125}{\second} after gait initiation with an accuracy of 80\% on a high frequency and low noise dataset.

There is still fewer research concerning intention detection of cyclists. In \cite{Pool.2017}, Pool et al. introduced a motion model for cyclist path prediction from a moving vehicle including knowledge of the local road topology. The authors were able to improve the prediction accuracy by incorporation of different motion models for canonical directions.

In our previous work \cite{Hubert.2017}, starting behavior of cyclists at an urban intersection was investigated and grouped into two different motion patterns. It was shown that 29\% of the cyclists initiate the starting motion with an arm movement. Furthermore, cyclists' heads moved on average \SI{0.33}{\second} earlier than the bike. A two-stage cooperative intention detection process for cyclists was introduced in \cite{Bieshaar.2017}. Depending on the detected movement primitives in the first stage, specialized models for forecasting future positions were weighted in the second stage. Thereby, a cooperation between smart devices and infrastructure-based sensors was used for the detection of starting motions. This approach stabilized the detection process and lowered the forecasting error of trajectories.

In this work, we use a deep residual CNN (ResNet) to detect starting motions with MHIs. In the past few years, CNNs have lead to tremendous progress in the field of image classification. The ResNet architecture was introduced by He et al. \cite{He2016DeepRL} and was used to win the 2015 ImageNet Large Scale Visual Recognition Challenge \cite{ILSVRC15}.

\subsection{Main Contributions and Outline of this Paper}

Our main contribution is a new method for early detection of cyclist starting motions in real world traffic scenarios. The approach uses a deep neural network with ResNet architecture. By combining MHIs with a ResNet classifier and performing a frame by frame classification we are able to safely detect starting motions within \SI{0.144}{\second}. We compare our approach to an existing method used to detect movement transitions of pedestrians using MCHOG and an SVM, which we adapted for cyclists. Both methods are evaluated in real world scenarios at an urban intersection with 394 starting scenes. The ResNet method outperforms the MCHOG approach in both robustness against false positives and detection time.

%This article describes two approaches for trajectory prediction of cyclists and compares them to a KF approach using a CV model. The first approach uses a physical model of the cyclists' velocity to predict their future positions. The second approach is based on polynomial least-squares approximations and multilayer perceptron ANN. The advantage of the ANN method is its independence from motion types such as "Starting", "Stopping", "Waiting" or "Passing". Both methods allow for a higher accuracy for predicting a cyclist's future position in comparison to the KF approach.

\section{\large Method}
\label{sec_method_overview}

This section outlines the two methods to detect movement transitions between \textit{waiting} and \textit{moving} phases of cyclists using MHIs. First, we describe how the dataset used to evaluate our algorithms is created in Sec. \ref{sec_data_acquisition_method}.
The generation of MHIs from image sequences is described in Sec. \ref{sec_method_mhi_generation}. Sec. \ref{sec_method_mchog} and Sec. \ref{sec_method_resnet} contain the methods for starting motion detection using MCHOG and ResNet. Finally, in Sec. \ref{sec_evaluation_method}, we present our evaluation method.

\begin{figure}
\begin{center}
	\includegraphics[width = 0.9\columnwidth]{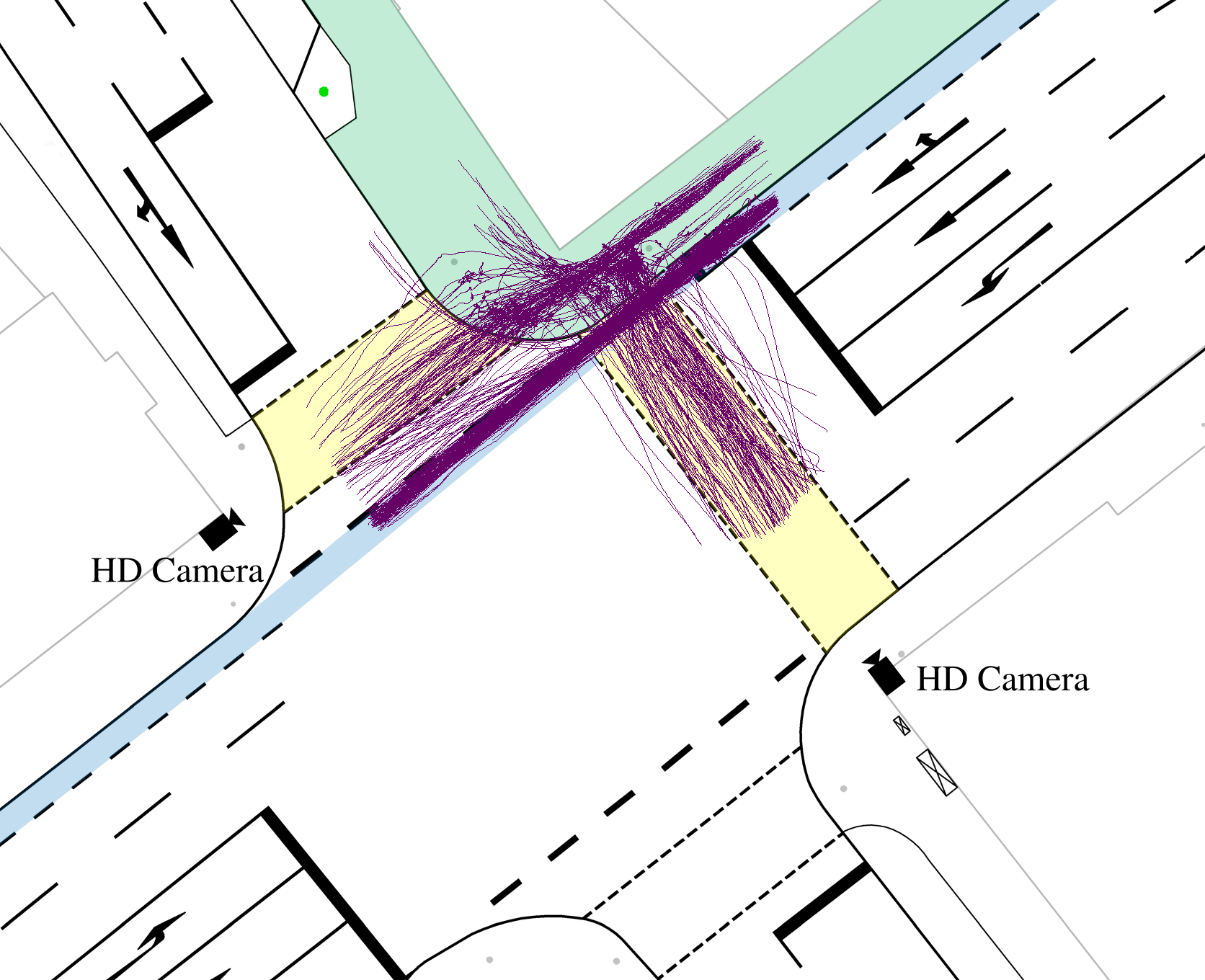}
	\caption{Overview of the intersection with all starting movements.}
	\label{fig:intersection}
\end{center}
\vskip -6mm
\end{figure}

\begin{figure*} [!h]
	\includegraphics[width = 2.0\columnwidth]{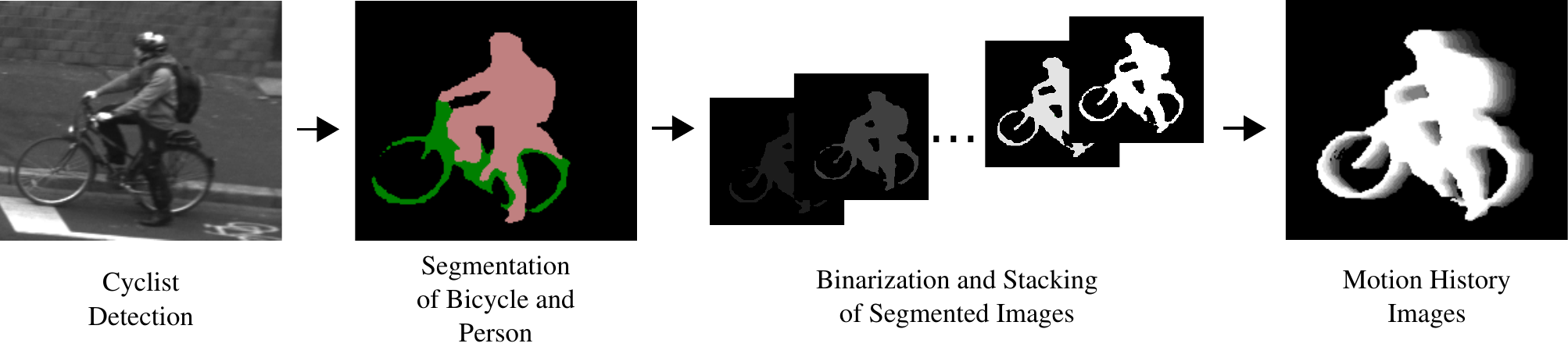}
	\vskip -3mm
	\caption{Generation of MHIs}
	\label{fig:mhi_generation}
	\vskip -3mm
\end{figure*}

\subsection{ Data Acquisition and Preprocessing}
\label{sec_data_acquisition_method}

To train and test the algorithms, we created a dataset containing 394 scenes of starting cyclists recorded at an urban intersection equipped with two HD cameras \cite{Goldhammer2012Inters}, with a frame rate of \SI{50}{\hertz}, arranged in a wide angle stereo camera system (Fig. \ref{fig:intersection}). The camera field of view covers a sidewalk (Fig. \ref{fig:intersection}, green) with two pedestrian crossings (yellow) and a bicycle lane (blue).

The dataset consists of 49 male and female test subjects, who were instructed to move between certain points on the intersection, while following the traffic rules, which lead to 89 starting motions. Additionally, 305 starting motions of uninstructed cyclists were recorded at random times, resulting in 394 starting motions total. The set was divided with a 60-20-20 split into training, validation, and test data. The trajectories of the recorded cyclists are shown in Fig. \ref{fig:intersection} in purple.

To generate the input of the classifiers $x_t$ containing the MHIs, the head positions were labeled manually in every image. 
A region of interest (ROI) with a size large enough to enclose the cyclist including the bicycle was chosen and based on the head position used to crop the images of the past $N$ time steps, which are used to create the MHI, as described in Sec. \ref{sec_method_mhi_generation}. The used ROI size is 192$\times$\SI{160}{\pixel}.

The output of the classifier $\hat{y}_t$ contains the class probabilities $P_{waiting}$ and $P_{moving}$. Additionally, an auxiliary class \textit{starting} for evaluation of the classifier is introduced. The labels were created manually and are defined as follows: An image is labeled \textit{waiting}, while neither the wheel of the bicycle is moving, nor the cyclist is performing a movement which leads to a starting motion. Every frame between the first visible movement of the cyclist that leads to a starting motion and the first movement of the bicycle wheel is labeled as \textit{starting}. Finally, every frame after the first movement of the bicycle wheel is labeled \textit{moving}. For training, \textit{starting} and \textit{moving} are merged into one class.

\subsection{ Generation of Motion History Images }
\label{sec_method_mhi_generation}

In this section, we describe how the MHIs used to classify the starting motions are generated. The generation is depicted in Fig. \ref{fig:mhi_generation}. In the first step, an ROI enclosing the detected cyclist and bicycle is created on the camera image which contains the side view of the cyclist (Fig. \ref{fig:mhi_generation}, left). The image is cropped to the size of the ROI and fed to a semantic segmentation. For the segmentation, the ResNet from \cite{zifeng.2016} pretrained on the CoCo dataset \cite{Lin2014} and trained on the PASCAL VOC dataset \cite{Everingham.2010} is used to assign classes to every pixel in the image. We use the VOC dataset over the Cityscapes dataset \cite{Cordts2016Cityscapes}, because it contains images of cyclists from different angles, which are very close to our stationary camera dataset, whereas the Cityscapes dataset consists solely of images recorded from a car. The segmentation outputs 20 classes of which the classes \textit{person}, \textit{bicycle}, and \textit{motorbike} are used to generate the silhouettes of cyclists and bicycles (Fig. \ref{fig:mhi_generation}, second from left). The class \textit{motorbike} is used, since parts of the bicycle are often misclassified as motorcycle. The image is binarized by setting these three classes to one and the other classes to zero. To generate the MHI (Fig. \ref{fig:mhi_generation}, right), binarized images $I(u,v,t)$ at different time steps $t$ are multiplied by a decay value $\tau(t)=\frac{N-t}{N}$, where $N$ is the number of past images used and $t$ is the $t^{th}$ image in the sequence, where $t=0$ is the most recent image. The decayed images are then stacked using Algorithm 1.

\begin{algorithm}
	\caption{MHI Generation}\label{euclid}
	\begin{algorithmic}[1]
		\State $I(u,v,t) \gets$ sequence of images with time step t and pixel positions $u$ and $v$, where $I(u,v,0)$ is the most recent image
		\State $N \gets$ number of time steps in $I(u,v,t)$
		\State $W \gets$ image width
		\State $H \gets$ image height
		\State $MHI(u, v):=0$
		\For{$t=N-1\text{ to } 0$}\Comment{iterate over all images, start with oldest}
		\State $\tau(t)=\frac{N-t}{N}$\Comment{calculate decay value $\tau$}
		\For{$u=0\text{ to } W-1$}\Comment{go through all pixels}
		\For{$v=0\text{ to } H-1$}
		\If{$I(u, v, t)==1$} \Comment{update MHI}
		\State $MHI(u, v)=\tau(t)\cdot I(u, v, t)$
		\EndIf
		\EndFor
		\EndFor		
		\EndFor
	\end{algorithmic}
\end{algorithm}

\subsection{MCHOG Detector}
\label{sec_method_mchog}

To detect cyclist starting motions using MCHOG, the method used to detect pedestrian motions described in \cite{KoehlerMag} is adapted to cyclists. 
The MCHOG descriptor is generated by computing the magnitude and orientation of the gradients in the MHI, dividing the image into cells and computing cell histograms. In contrast to the original implementation of the HOG descriptor \cite{Dalal.2005}, a block normalization of cells is not performed, as it reduces the local differences between neighboring cells.
The concatenated cell histograms result in the MCHOG descriptor, which is used as input of a linear SVM classifier.
The HOG descriptors are computed on MHIs. To reduce the number of features, the MHIs are resized to 128$\times$\SI{96}{\pixel}. To generate probability outputs from the SVM, probability calibration was performed using Platt's algorithm \cite{Platt.1999}.

\subsection{Deep Residual Network Detector}
\label{sec_method_resnet}

\begin{figure} [!hb]
	\vskip -5mm
	\includegraphics[width = 0.9\columnwidth]{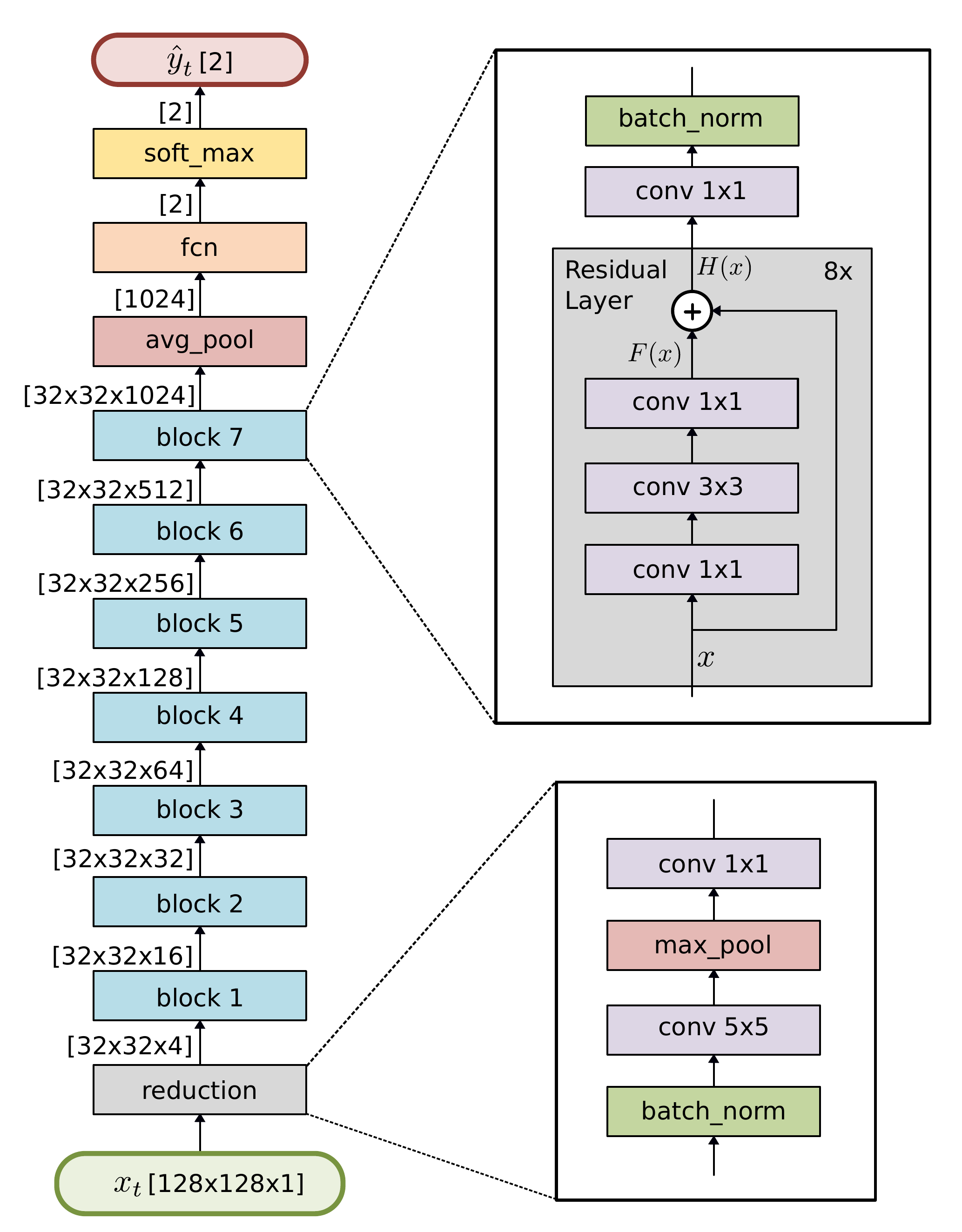}
	\caption{ResNet architecture with a reduction layer (bottom right) and residual blocks (top right).}
	\label{fig:resnet_archtiecture}
\end{figure}

In this section, we describe the detection of starting motions using a ResNet architecture which was first introduced by He et al. in \cite{He2016DeepRL}. The authors showed, that their network is easier to train and generates a higher accuracy compared to conventional CNNs, by addressing a degradation problem, where adding more layers to conventional deep models leads to a higher training error. They introduced residual building blocks (Fig. \ref{fig:resnet_archtiecture}, upper right), where instead of directly modeling a target function $H(x)$ using a non linear model $F(x)$, they created a bypass and added the input to the output $F(x)+x$. Thus $F(x)$ models the residual of $H(x)$. One explanation for the degradation problem is that it is difficult to model identities with non linear layers. Using a residual layer, the identity can be modeled by driving the weights to zero.
By stacking residual blocks, the authors were able to train a network with 152 layers, which produced substantially better results than shallower networks.

Our network architecture, which is described in Fig. \ref{fig:resnet_archtiecture}, is based on the architecture in \cite{He2016DeepRL}.
The MHI, resized to 128$\times$\SI{128}{\pixel}, is used as input where a reduction layer (Fig. \ref{fig:resnet_archtiecture}, lower right) consisting of batch normalization, a 5$\times$5 convolution, and a max pooling layer is used to reduce the image dimension. A 1$\times$1 convolution is applied to generate four feature maps. The feature maps are then passed to a residual block, which is described in Fig. \ref{fig:resnet_archtiecture} on the upper right. In our network, a residual block consists of eight residual layers with bottleneck architecture to reduce the computational effort, followed by a 1$\times$1 convolution to generate the output feature maps and a batch normalization layer. After seven residual blocks an average pooling is applied to receive a feature vector containing 1024 features, which are classified by a fully connected layer (fcn) with softmax activation to generate probabilities. To speed up the training process, batch normalization layers are added at the network input and after every residual block.

%\subsection{Trajectory Prediction using a Constant Velocity Model}
%\label{sec_method_cv}

%In this section we describe the prediction of cyclists' future positions by using a KF with Constant Velocity (CV) model. For the prediction, cyclists' positions from past time steps were used to update the KF. To predict the future position $\vec{x}_{pred}$, the transition matrix $A$ with the prediction time $t_{pred}$ and the state vector at the current time $\vec{x}_{t_c}$ with the position $(x; y)$ and the velocity $(\dot x; \dot y)$ were used in Eq. \ref{eq:KF_predict}.

%\begin{equation}
%A=\left( \begin{matrix}
%1 & 0 & t_{pred} & 0 \\
%0 & 1 & 0 & t_{pred} \\
%0 & 0 & 1 & 0 \\
%0 & 0 & 0 & 1
%\end{matrix} \right) ; \vec{x}= \left( \begin{matrix}  x \\ y \\ \dot x \\ \dot y \end{matrix} \right); 
%\label{eq:cv_model}
%\end{equation}

%\begin{equation}
%\vec{x}_{pred} = A \cdot \vec{x}_{t_c}
%\label{eq:KF_predict}
%\end{equation}

\subsection{ Evaluation Method }
\label{sec_evaluation_method}

To evaluate our algorithms, we used the recorded dataset described in Sec. \ref{sec_data_acquisition_method}, i.e., the evaluation was done offline.

\begin{figure}
\begin{center}
	\includegraphics[width = 0.9\columnwidth]{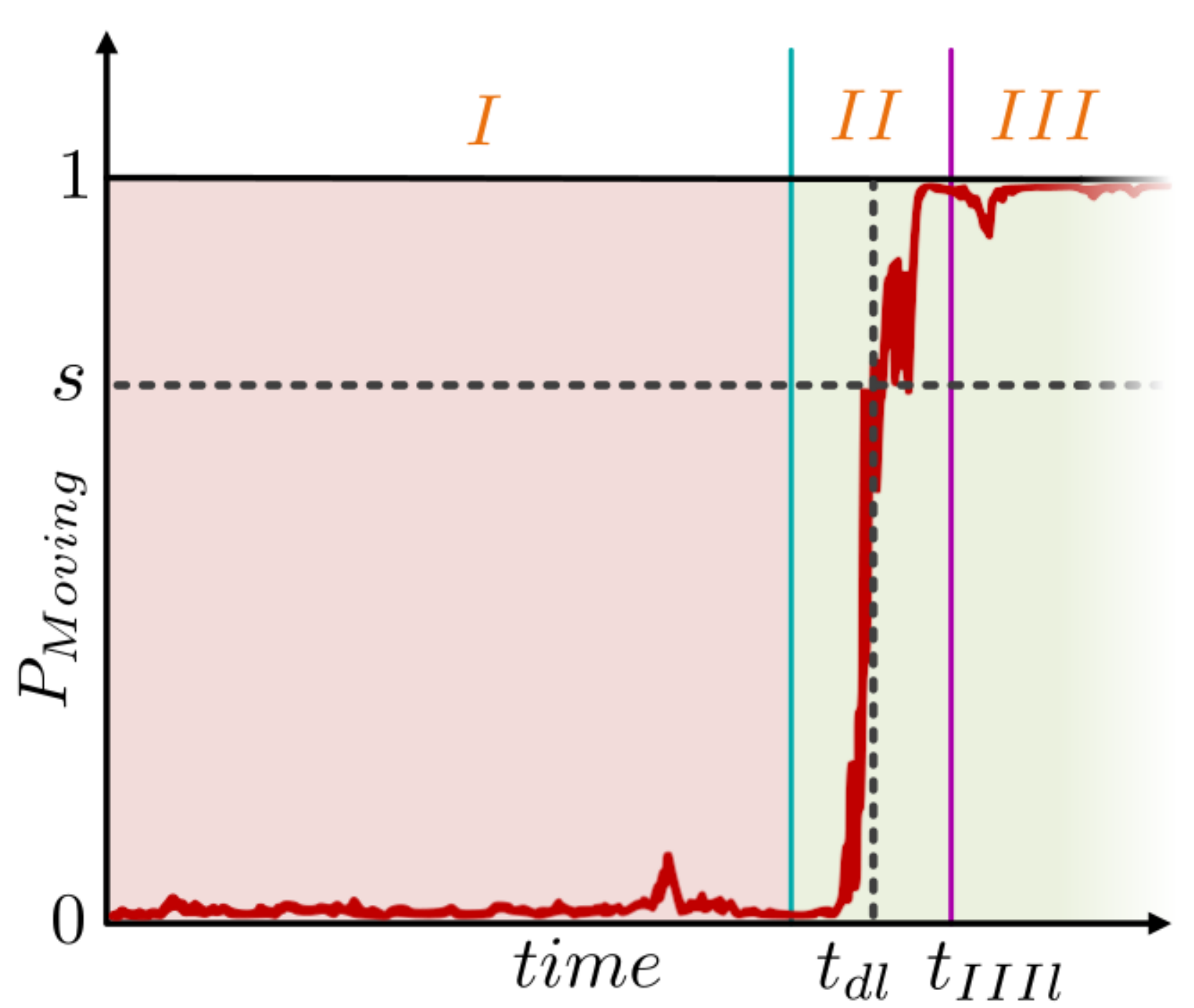}
	\vskip -3mm
	\caption{Exemplary classification output of a scene, with moving probability $P_{Moving}$ (red), labeled \textit{starting} time (blue), and labeled \textit{moving} time $t_{IIIl}$ (purple). A chosen threshold $s$, leads to detection time $t_{dl}$.}
	\label{fig:example_classification}
\end{center}
\end{figure}

The performance of both detectors was determined by a scene wise evaluation, where one scene starts after the cyclists stopped and ends when the cyclist leaves the field of view of the side view camera. Fig. \ref{fig:example_classification} shows an exemplary output of a scene, where $P_{Moving}$ (red line) is plotted over time. Phase $I$ is the \textit{waiting} phase, phase $II$ and $III$ are \textit{starting} and \textit{moving} phases, respectively. A desired output of the detector is shown in Fig. \ref{fig:example_classification}, $P_{Moving}$ maintains a low value during Phase $I$, increases in phase $II$, and remains at a high level during phase $III$. A starting movement is detected when $P_{moving}$ reaches a certain threshold ($s$ in Fig. \ref{fig:example_classification}). A scene is rated as true positive, if the threshold is reached in phase $II$ or $III$. If the threshold is reached during phase $I$, the scene is rated as false positive. If the threshold is never reached, it is rated as false negative. We do not consider true negatives, since every scene results in \textit{moving}.

Using this method, we calculate the $precision$ and the $F_{1}$-score for thresholds between zero and one with a step size of 0.02. Additionally, we evaluate the detection time by calculating the mean time difference $\overline{\delta_{t}}$ between the detection time $t_{dl}$ and the start time of phase $III$ $t_{IIIl}$ in the $l^{th}$ sequence of all true positives over all $L$ sequences (Eq. \ref{eq:mean_detection_time}), where smaller values indicate faster detection. 

\begin{equation}
	\overline{\delta_{t}}=\frac{1}{L}\cdot\sum_{l=1}^{L}(t_{dl}-t_{IIIl})
	\label{eq:mean_detection_time}
\end{equation}

\section{Experimental results}
\label{sec_ResultsOutline}

This section describes the evaluation of the proposed methods and compares their results.

\begin{figure*} 
	\begin{center}
		\includegraphics[width = 1.9\columnwidth]{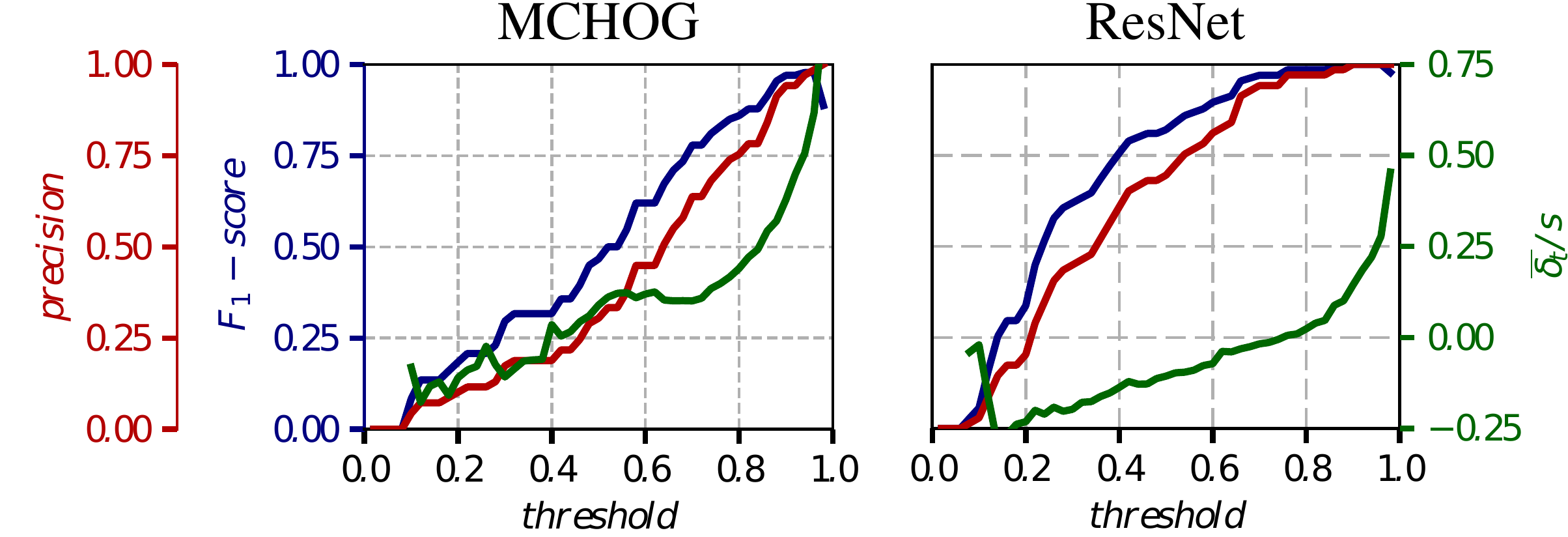}
		\caption{$F_1$-scores (blue), $precision$ (red), and mean detection time $\overline{\delta_t}$ over probability thresholds of MCHOG (left) and ResNet (right).}
		\label{fig:mchog_f1}
	\end{center}
	\vskip -5mm
\end{figure*}

\begin{table}
	\begin{center}
	\caption{Values used in MCHOG parameter sweep.}
	\scalebox{0.85}{
		\begin{tabular}{ | l | c | c | c | c |}
			\hline
			Param. & cell size x & cell size y & nbins & $C$ param \\ \hline
			Value & $\{8, 16, 32\}$ & $\{8, 16, 32\}$ & $\{6, 12, 16\}$ & $\{2^c|-9<c<5:c\in\mathbb{N}\}$ \\
			\hline
		\end{tabular}}
		\label{tab:paramsweep_mchog}
	\end{center}
	\vskip -8mm
\end{table}

\subsection{MCHOG Results}
\label{sec_results_mchog}

In this section, we present the results of the detection using MCHOG in combination with an SVM. We performed a grid search over the cell size in x and y direction, the number of bins in a histogram and the $C$ parameter of the SVM. The values used in the parameter sweep can be found in Tab.~\ref{tab:paramsweep_mchog}. The dataset described in Sec. \ref{sec_data_acquisition_method} was used for training, validation and test.

We generated the $F_1$-score and the mean detection time $\overline{\delta_t}$ needed to achieve the highest $F_1$-score for every parameter configuration using the validation set. The five best and three worst validation results are shown in Tab.~\ref{tab:results_paramsweep_mchog}. It shows that four of the five best results have the same MCHOG parameters and only differ in the $C$ parameter of the SVM. Furthermore, the parameter sweep yielded 51 detectors that reached an $F_1$-score of 100\%, where $\overline{\delta_t}$ ranges from \SI{0.565}{\second} for the fastest detector to $1.11\,s$ for the slowest detector. The classifiers with large cell size in y direction and low number of histogram bins yielded the lowest $F_1$-scores.

\begin{table}
	\begin{center}
	\caption{Validation results from MCHOG parameter sweep.}
	\scalebox{1.0}{
		\begin{tabular}{ | c | c || c | c | c | c |}
			\hline
			 $F_1$ & $\overline{\delta_t}$ & cell size x & cell size y & nbins & $C$ param \\ \hline \hline
			1.0 &\SI{0.565}{\second} & 32 & 8 & 18 & 0.03125 \\ \hline
			1.0 &\SI{0.578}{\second} & 32 & 8 & 18 & 0.0625 \\ \hline
			1.0 &\SI{0.586}{\second} & 32 & 8 & 18 & 0.125 \\ \hline
			1.0 &\SI{0.608}{\second} & 8 & 8 & 12 & 0.25 \\ \hline
			1.0 &\SI{0.609}{\second} & 32 & 8 & 18 & 0.25 \\ \hline
			\rotatebox{90}{... } & \rotatebox{90}{... } & \rotatebox{90}{... } & \rotatebox{90}{... } & \rotatebox{90}{... } & \rotatebox{90}{... } \\ \hline
			0.915 &\SI{0.968}{\second} & 32 & 32 & 6 & 4 \\ \hline
			0.915 &\SI{0.968}{\second} & 32 & 32 & 6 & 2 \\ \hline
			0.915 &\SI{0.968}{\second} & 32 & 32 & 6 & 8 \\
			\hline
		\end{tabular}}
		\label{tab:results_paramsweep_mchog}
	\end{center}
	\vskip -5mm
\end{table}

\parskip = 0pt

To generate the test results, the detector with pareto-optimal validation scores was chosen, i.e., greatest $F_1$-score and lowest $\overline{\delta_t}$. Fig. \ref{fig:mchog_f1} (left) shows the overall results of the detector. To generate the plot, the $F_1$-score, the $precision$ and $\overline{\delta_t}$ were generated for different probability thresholds (as described in Sec. \ref{sec_evaluation_method}) and plotted over thresholds from zero to one. Our evaluation shows that the detector reaches an $F_1$-score of 90\% \SI{0.274}{\second} after the first movement of the bicycle wheel and the highest $F_1$-score of 97.8\% is reached after \SI{0.506}{\second}. 

\begin{figure} 
	\begin{center}
		\includegraphics[width = 0.75\columnwidth]{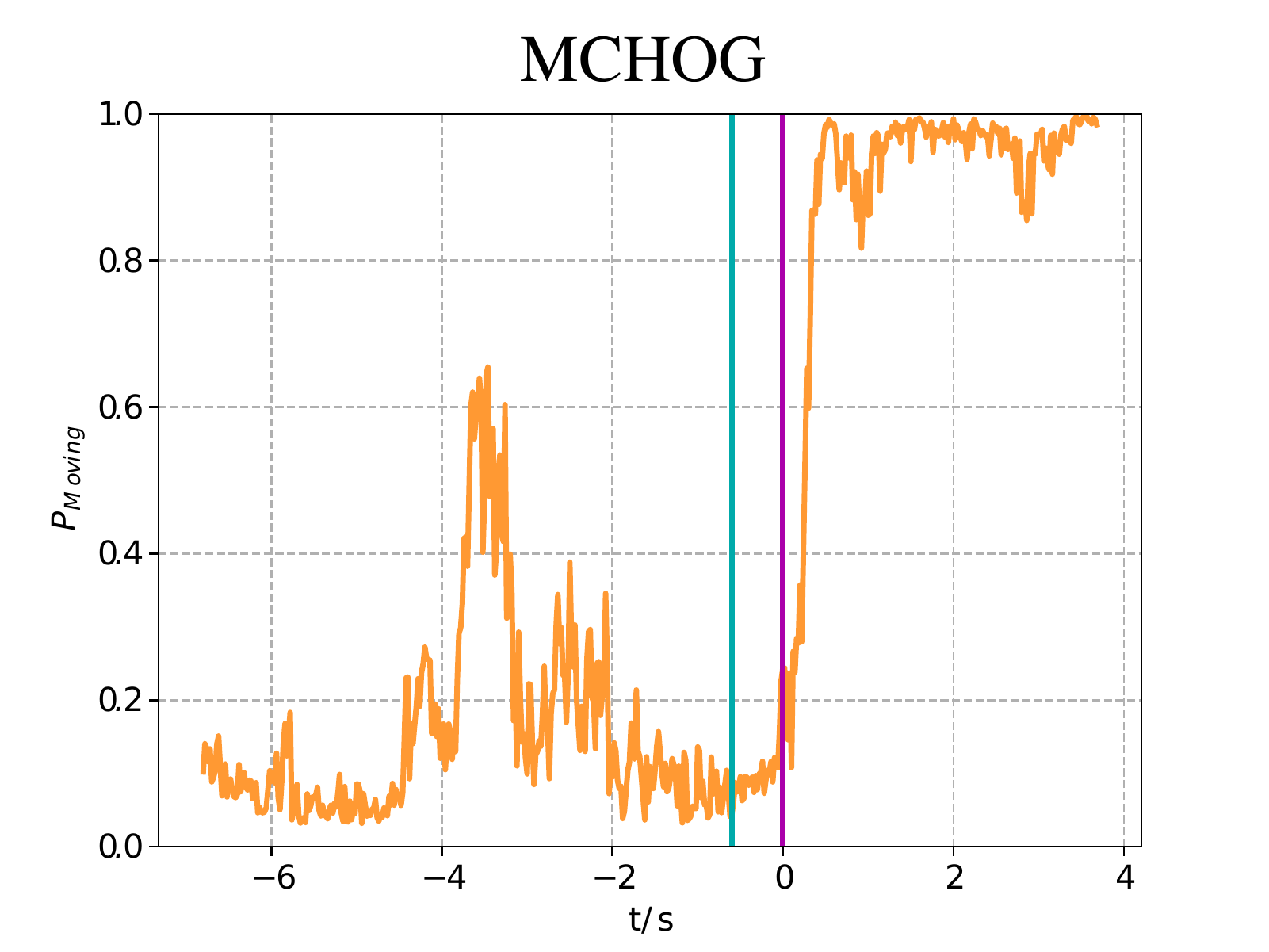}
		\includegraphics[width = 0.75\columnwidth]{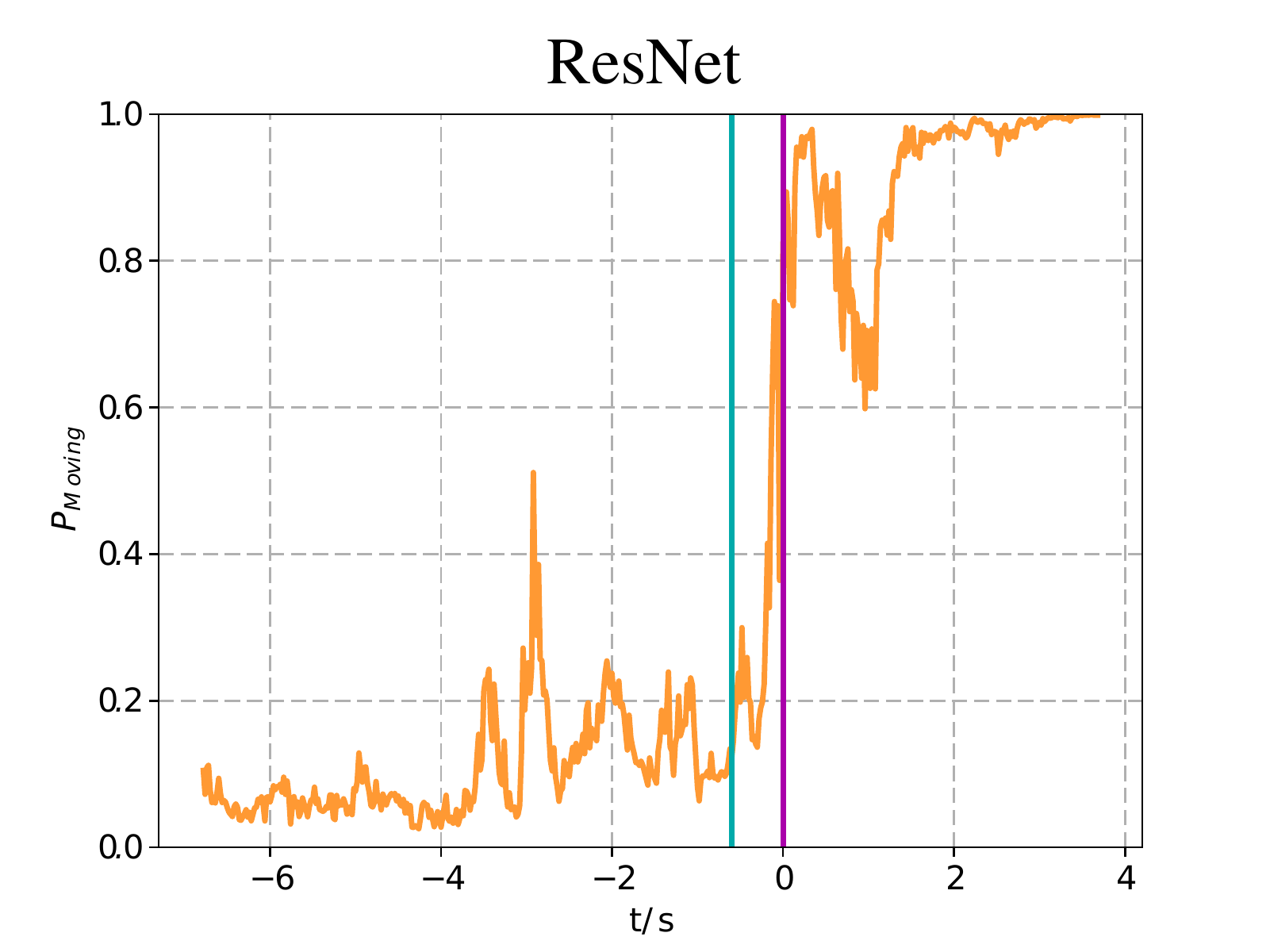}
		\caption{Example detection 1: Moving probabilities of MCHOG (top) and ResNet (bottom) detection with crossing pedestrian between \SI{-5}{\second} and \SI{-2}{\second}. }
		\label{fig:ex1_classification}
	\end{center}
	\vskip -5mm
\end{figure}

\begin{figure}
	\includegraphics[width = 0.44\columnwidth]{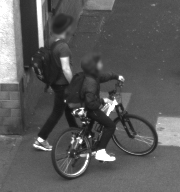}
	\includegraphics[width = 0.55\columnwidth]{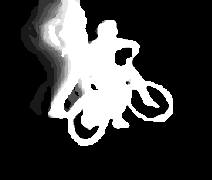}
	\caption{Example detection 1: Pedestrian passing behind cyclist.}
	\label{fig:ex1_image}
\end{figure}

\begin{figure}
	\begin{center}
		\includegraphics[width = 0.75\columnwidth]{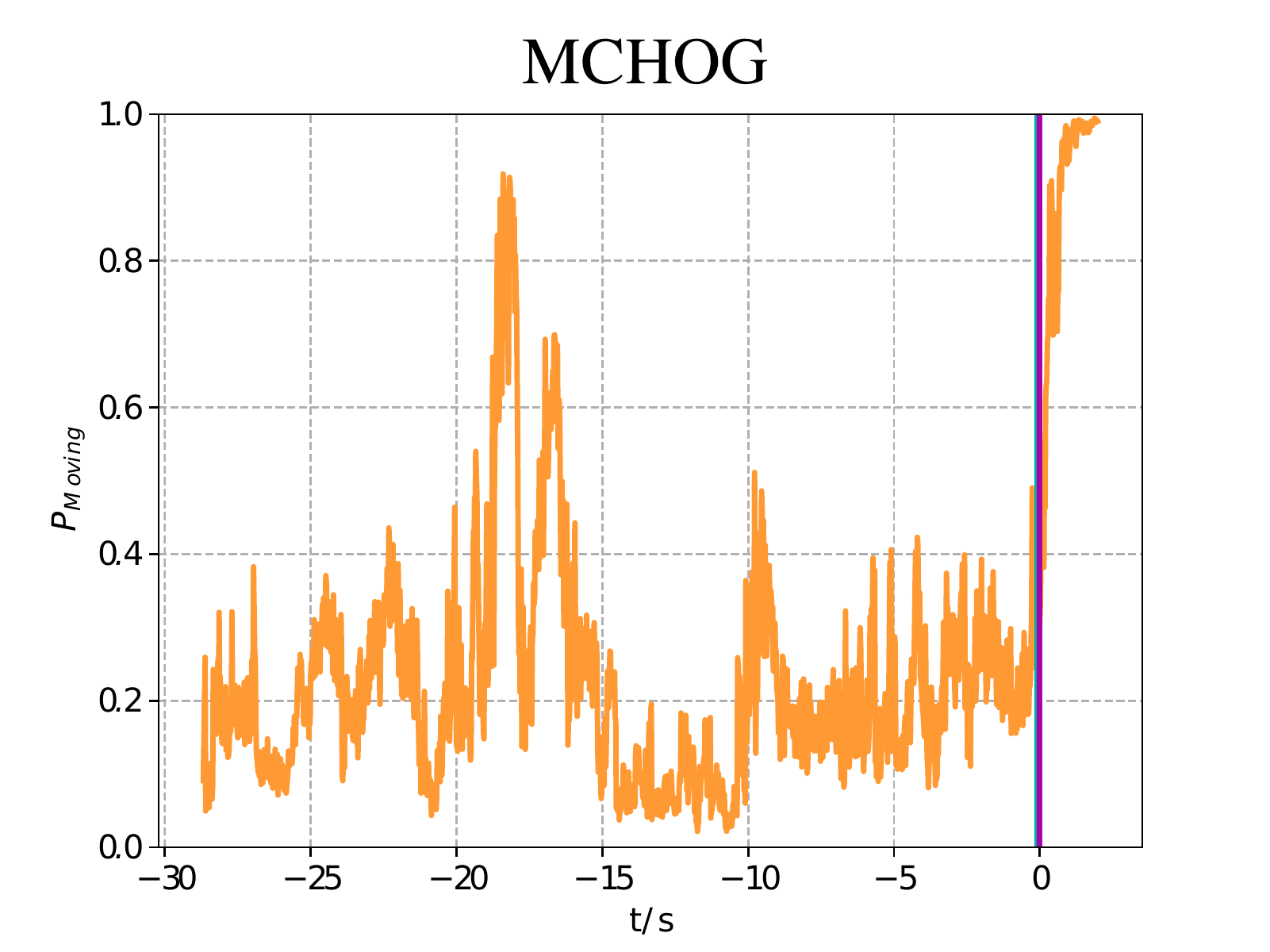}
		\includegraphics[width = 0.75\columnwidth]{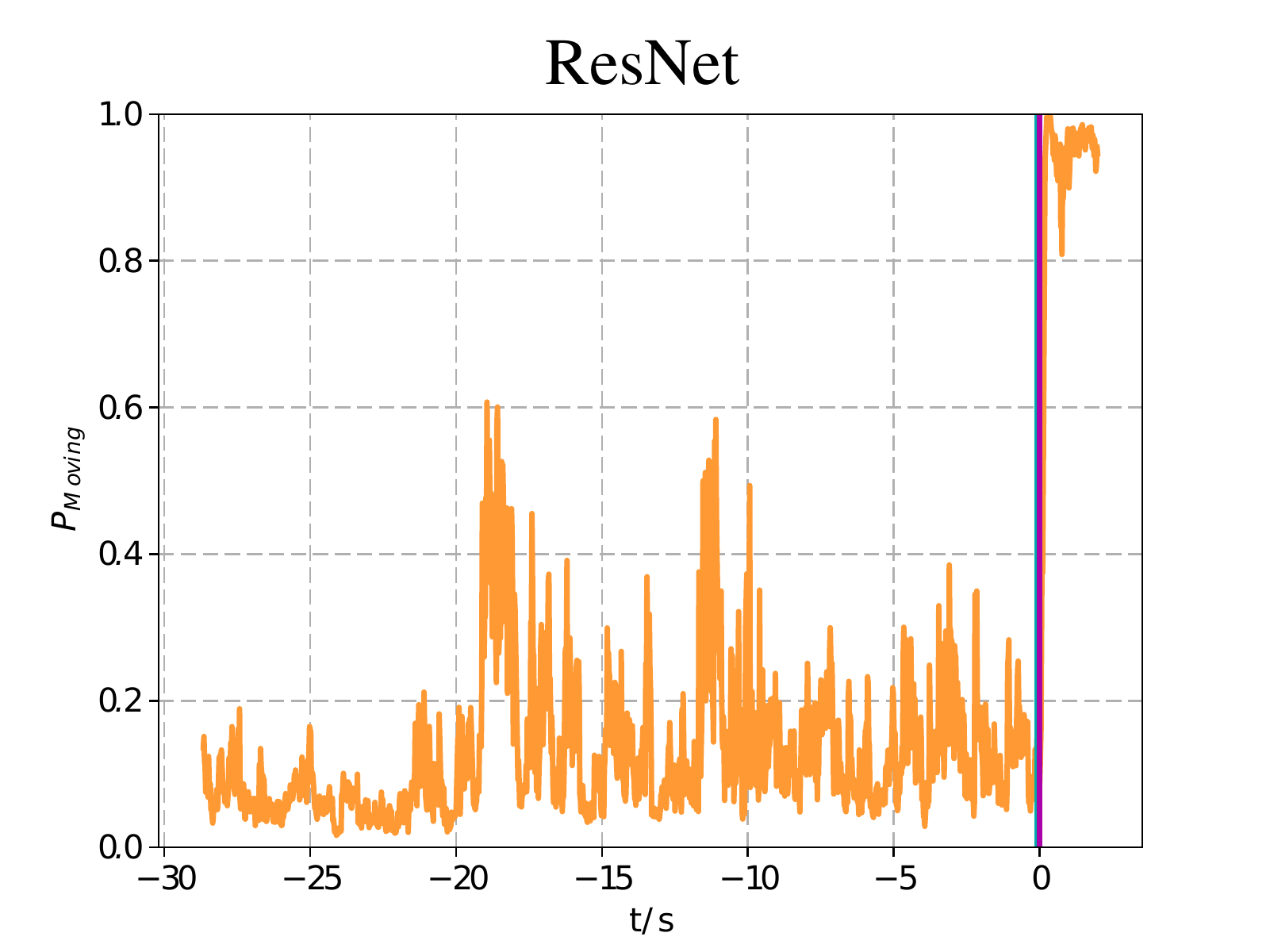}
		\caption{Example detection 2: Moving probabilities of MCHOG (top) and ResNet (bottom) detection with crossing pedestrian between \SI{-20}{\second} and \SI{-15}{\second}.}
		\label{fig:ex2_classification}
	\end{center}
\end{figure}

\begin{figure}
	\includegraphics[width = 0.445\columnwidth]{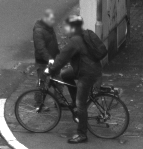}
	\includegraphics[width = 0.545\columnwidth]{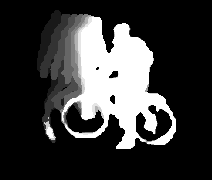}
	\caption{Example detection 2: Pedestrian passing close behind cyclist.}
	\label{fig:ex2_image}
	\vskip -5mm
\end{figure}

The classifier is robust against movements of the cyclist that do not lead to starting motion, however strongly reacts to movements of pedestrians passing in the background. Fig. \ref{fig:ex1_classification} and Fig. \ref{fig:ex2_classification} show two example classifications with pedestrians moving in the background of the waiting cyclists, which leads to an increase in $P_{Moving}$. Fig. \ref{fig:ex1_image} and \ref{fig:ex2_image} show the passing pedestrians in the camera image and the corresponding MHI. The peak in $P_{Moving}$ is reached when the pedestrian is occluded by the cyclist and only the motion contour of the pedestrian is visible, making it appear that the motion contour belongs to the cyclist. The second peak in Fig. \ref{fig:ex2_classification} between  \SI{-8}{\second} and \SI{-11}{\second} results from a strong forward movement of the cyclist and the bicycle.

%\begin{figure} [h]
%	\begin{center}
%		\includegraphics[width = 0.99\columnwidth]{images/f1_mchog_head.pdf}
%		\label{fig:resnet_f1}
%		\vskip 2mm
%		\includegraphics[width = 0.99\columnwidth]{images/f1_resnet_head.pdf}
%		\caption{vorl\"aufiges Ergebnis uninstruierte Radfahrer}
%		\label{fig:mchog_f1}
%	\end{center}
%\end{figure}

\subsection{Residual Network Results}
\label{sec_classification_results}

The ResNet detector was trained using the same dataset as the MCHOG classifier. As optimizer we used RMSProp in combination with a cross entropy loss function and a batch size of 10. The training was executed on an NVIDIA GTX 1080 Ti GPU using Tensorflow \cite{tensorflow2015}. The network was trained for 120,000 iterations.

To choose the best network, a validation step was performed every 250 iterations, where the  $F_1$-score and $\overline{\delta_t}$ were calculated for the validation set. The network with the best validation scores reaches an $F_1$-score of 100\% with $\overline{\delta_t}=$\SI{0.175}{\second} at iteration 66,000 and was used to create the test results.

The overall results are shown in Fig. \ref{fig:mchog_f1} (right). The classifier reaches an $F_1$-score score of 90$\%$ after \SI{-0.038}{\second} and the highest $F_1$-score of 100\% is reached after \SI{0.144}{\second}.

Like the MCHOG classifier, the ResNet is not influenced by small movements of the cyclist during the \textit{waiting} phase. Scenes with pedestrians passing in the background of the cyclists result in an increase of $P_{Moving}$, however, compared to the MCHOG, the ResNet does not react as significantly. Additionally, the ResNet outperforms the MCHOG when it comes to detection time. Concerning the detectors with the best $F_1$-scores, the ResNet is able to detect starting motions \SI{0.362}{\second} earlier on average, compared to the MCHOG.

\section{\large Conclusions and Future Work}
\label{sec_conclusion}

In this article, we presented two methods based on MHIs to detect starting motions of cyclists. The methods were tested in real world scenarios at an urban intersection. 
We adapted an existing method, which uses MCHOG descriptors and an SVM, to detect motions of pedestrians to cyclists and presented a new approach by using ResNet to detect starting motions. 

Using the MCHOG, we achieve an $F_1$-score of 97.8\% after \SI{0.506}{\second}. The ResNet approach outperforms the MCHOG in both robustness against false positives and detection time with a maximum $F_1$-score of 100\% after \SI{0.144}{\second} on average. 

Our future work will focus on how the developed methods can be used to further improve trajectories forecast algorithms. We also intend to adapt our method to moving vehicles.
Furthermore, we will investigate how the methods can be utilized in a cooperative way between different traffic participants to generate a comprehensive model of the environment.

\section{\large Acknowledgment}

This work results from the project DeCoInt$^2$, supported by the German Research Foundation (DFG) within the priority program SPP 1835: ``Kooperativ interagierende Automobile'', grant numbers DO 1186/1-1 and SI 674/11-1. Additionally, the work is supported by ``Zentrum Digitalisierung Bayern''.

% trigger a \newpage just before the given reference
% number - used to balance the columns on the last page
% adjust value as needed - may need to be readjusted if
% the document is modified later
%\IEEEtriggeratref{8}
% The "triggered" command can be changed if desired:
%\IEEEtriggercmd{\enlargethispage{-5in}}

% references section

% can use a bibliography generated by BibTeX as a .bbl file
% BibTeX documentation can be easily obtained at:
% http://www.ctan.org/tex-archive/biblio/bibtex/contrib/doc/
% The IEEEtran BibTeX style support page is at:
% http://www.michaelshell.org/tex/ieeetran/bibtex/
\bibliographystyle{IEEEtran}
% argument is your BibTeX string definitions and bibliography database(s)
%\bibliography{IEEEabrv,../bib/paper}
%
% <OR> manually copy in the resultant .bbl file
% set second argument of \begin to the number of references
% (used to reserve space for the reference number labels box)

{\small
\bibliography{egbib,sk,vk}

% Generated by IEEEtran.bst, version: 1.13 (2008/09/30)
\begin{thebibliography}{10}
\providecommand{\url}[1]{#1}
\csname url@samestyle\endcsname
\providecommand{\newblock}{\relax}
\providecommand{\bibinfo}[2]{#2}
\providecommand{\BIBentrySTDinterwordspacing}{\spaceskip=0pt\relax}
\providecommand{\BIBentryALTinterwordstretchfactor}{4}
\providecommand{\BIBentryALTinterwordspacing}{\spaceskip=\fontdimen2\font plus
\BIBentryALTinterwordstretchfactor\fontdimen3\font minus
  \fontdimen4\font\relax}
\providecommand{\BIBforeignlanguage}[2]{{%
\expandafter\ifx\csname l@#1\endcsname\relax
\typeout{** WARNING: IEEEtran.bst: No hyphenation pattern has been}%
\typeout{** loaded for the language `#1'. Using the pattern for}%
\typeout{** the default language instead.}%
\else
\language=\csname l@#1\endcsname
\fi
#2}}
\providecommand{\BIBdecl}{\relax}
\BIBdecl

\bibitem{WHO.2015}
\BIBentryALTinterwordspacing
{World Health Organization}, ``{G}lobal {S}tatus {R}eport on {R}oad {S}afety
  2015,'' 2015. [Online]. Available:
  \url{http://www.who.int/violence_injury_prevention/road_safety_status/2015/en/}
\BIBentrySTDinterwordspacing

\bibitem{Keller.2011}
C.~Keller, C.~Hermes, and D.~Gavrila, ``Will the pedestrian cross?
  probabilistic path prediction based on learned motion features,'' in
  \emph{Pattern Recognition}.\hskip 1em plus 0.5em minus 0.4em\relax Springer,
  2011, vol. 6835, pp. 386--395.

\bibitem{Bieshaar.2017}
M.~Bieshaar, S.~Zernetsch, M.~Depping, B.~Sick, and K.~Doll, ``Cooperative
  starting intention detection of cyclists based on smart devices and
  infrastructure,'' in \emph{International Conference on Intelligent
  Transportation Systems (ITSC)}, Yokohama, 2017.

\bibitem{KoehlerMag}
S.~Koehler, M.~Goldhammer, S.~Bauer, S.~Zecha, K.~Doll, U.~Brunsmann, and
  K.~Dietmayer, ``Stationary detection of the pedestrian's intention at
  intersections,'' \emph{IEEE Intelligent Transportation Systems Magazine},
  vol.~5, no.~4, pp. 87--99, 2013.

\bibitem{Quintero.2017}
R.~Quintero, I.~Parra, J.~Lorenzo, D.~Fernández-Llorca, and M.~A. Sotelo,
  ``Pedestrian intention recognition by means of a hidden markov model and body
  language,'' in \emph{International Conference on Intelligent Transportation
  Systems (ITSC)}, Yokohama, 2017.

\bibitem{Keller.2014}
C.~Keller and D.~Gavrila, ``Will the pedestrian cross? a study on pedestrian
  path prediction,'' \emph{Intelligent Transportation Systems, IEEE
  Transactions on}, vol.~15, no.~2, pp. 494--506, 2014.

\bibitem{Koehler2015}
S.~Koehler, M.~Goldhammer, K.~Zindler, K.~Doll, and K.~Dietmayer,
  ``Stereo-vision-based pedestrian's intention detection in a moving vehicle,''
  in \emph{International Conference on Intelligent Transportation Systems
  (ITSC)}, Gran Canaria, Sept 2015, pp. 2317--2322.

\bibitem{Quintero.2015}
R.~Quintero, I.~Parra, D.~F. Llorca, and M.~A. Sotelo, ``Pedestrian intention
  and pose prediction through dynamical models and behaviour classification,''
  in \emph{International Conference on Intelligent Transportation Systems
  (ITSC)}, Anchorage, 2015, pp. 83--88.

\bibitem{Pool.2017}
E.~A.~I. Pool, J.~F.~P. Kooij, and D.~M. Gavrila, ``Using road topology to
  improve cyclist path prediction,'' in \emph{IEEE Intelligent Vehicles
  Symposium (IV)}, Redondo Beach, June 2017, pp. 289--296.

\bibitem{Hubert.2017}
A.~Hubert, S.~Zernetsch, K.~Doll, and B.~Sick, ``Cyclists' starting behavior at
  intersections,'' in \emph{Intelligent Vehicles Symposium (IV)}, June 2017,
  pp. 1071--1077.

\bibitem{He2016DeepRL}
K.~He, X.~Zhang, S.~Ren, and J.~Sun, ``Deep residual learning for image
  recognition,'' in \emph{IEEE Conference on Computer Vision and Pattern
  Recognition (CVPR)}, Las Vegas, 2016, pp. 770--778.

\bibitem{ILSVRC15}
O.~Russakovsky, J.~Deng, H.~Su, J.~Krause, S.~Satheesh, S.~Ma, Z.~Huang,
  A.~Karpathy, A.~Khosla, M.~Bernstein, A.~C. Berg, and L.~Fei-Fei, ``{ImageNet
  Large Scale Visual Recognition Challenge},'' \emph{International Journal of
  Computer Vision (IJCV)}, vol. 115, no.~3, pp. 211--252, 2015.

\bibitem{Goldhammer2012Inters}
M.~Goldhammer, E.~Strigel, D.~Meissner, U.~Brunsmann, K.~Doll, and
  K.~Dietmayer, ``Cooperative multi sensor network for traffic safety
  applications at intersections,'' in \emph{International Conference on
  Intelligent Transportation Systems (ITSC)}, Anchorage, 2012, pp. 1178--1183.

\bibitem{zifeng.2016}
Z.~Wu, C.~Shen, and A.~van~den Hengel, ``Wider or deeper: {R}evisiting the
  resnet model for visual recognition,'' arXiv:1611.10080, 2016.

\bibitem{Lin2014}
T.-Y. Lin, M.~Maire, S.~Belongie, J.~Hays, P.~Perona, D.~Ramanan,
  P.~Doll{\'a}r, and C.~L. Zitnick, \emph{Microsoft COCO: Common Objects in
  Context}.\hskip 1em plus 0.5em minus 0.4em\relax Cham: Springer International
  Publishing, 2014, pp. 740--755.

\bibitem{Everingham.2010}
M.~Everingham, L.~Van~Gool, C.~K.~I. Williams, J.~Winn, and A.~Zisserman, ``The
  pascal visual object classes (voc) challenge,'' \emph{International Journal
  of Computer Vision}, vol.~88, no.~2, pp. 303--338, 2010.

\bibitem{Cordts2016Cityscapes}
M.~Cordts, M.~Omran, S.~Ramos, T.~Rehfeld, M.~Enzweiler, R.~Benenson,
  U.~Franke, S.~Roth, and B.~Schiele, ``The cityscapes dataset for semantic
  urban scene understanding,'' in \emph{Proc. of the IEEE Conference on
  Computer Vision and Pattern Recognition (CVPR)}, Las Vegas, 2016.

\bibitem{Dalal.2005}
N.~Dalal and B.~Triggs, ``{H}istograms of {O}riented {G}radients for {H}uman
  {D}etection,'' in \emph{{IEEE} {C}onference on {C}omputer {V}ision and
  {P}attern {R}ecognition (CVPR)}, San Diego, 2005, pp. 886--893.

\bibitem{Platt.1999}
J.~C. Platt, ``Probabilistic outputs for support vector machines and
  comparisons to regularized likelihood methods,'' in \emph{Advances in Large
  Margin Classifiers}.\hskip 1em plus 0.5em minus 0.4em\relax MIT Press, 1999,
  pp. 61--74.

\bibitem{tensorflow2015}
\BIBentryALTinterwordspacing
M.~Abadi \emph{et~al.}, ``{TensorFlow}: Large-scale machine learning on
  heterogeneous systems,'' 2015, software available from tensorflow.org.
  [Online]. Available: \url{https://www.tensorflow.org/}
\BIBentrySTDinterwordspacing

\end{thebibliography}
}

% that's all folks
\end{document}